\documentclass[lettersize,journal]{IEEEtran}
\usepackage{amsmath,amsfonts}
\usepackage{algorithmic}
\usepackage{algorithm}
\usepackage{array}
\usepackage[caption=false,font=normalsize,labelfont=sf,textfont=sf]{subfig}
\usepackage{textcomp}
\usepackage{stfloats}
\usepackage{url}
\usepackage{verbatim}
\usepackage{graphicx}
\usepackage{cite}
\usepackage{threeparttable}
\usepackage{xcolor}
\usepackage{pifont}
\usepackage{multirow}
\usepackage{colortbl}
\definecolor{mygray}{gray}{.9}
\definecolor{mypink}{rgb}{.99,.91,.95}
\definecolor{mycyan}{cmyk}{.3,0,0,0}
\definecolor{cvprblue}{rgb}{0.21,0.49,0.74}
\usepackage[pagebackref,breaklinks,colorlinks,citecolor=cvprblue]{hyperref}
\usepackage{bbding}
\usepackage{amsmath}  
\usepackage{csquotes}

\begin{document}

\title{Repurposing CLIP to Localize at Pixel Level}

\author{Jiaxiang Fang,~\IEEEmembership{Member,~IEEE,} Shiqiang Ma,~\IEEEmembership{Member,~IEEE,} Jing Wang, Siyu Chen, \\Fei Guo,~\IEEEmembership{Member,~IEEE,} and Shengfeng He,~\IEEEmembership{Senior Member,~IEEE}
\thanks{This work is supported by grants from the National Natural Science Foundation of China (Grants No. 62322215, 62532017 and 62402488), Natural Science Foundation of Hunan Province (Grants No. 2026JJ30018), the Guangdong Natural Science Funds for Distinguished Young Scholars (Grant 2023B1515020097), the National Research Foundation Singapore under the AI Singapore Programme (AISG Award No: AISG4-TC-2025-018-SGKR), and the Lee Kong Chian Fellowships. (Jiaxiang Fang and Shiqiang Ma contributed equally to this work.) (Corresponding authors: Fei Guo; Shengfeng He.)}
\thanks{Jiaxiang Fang is with the School of Computer Science and Engineering, Central South University, Changsha 410083, China, and is with the Advanced Technology Center Beijing AI Laboratory, Chao-Yang District, Beijing 100027, China (e-mail: 254701041@csu.edu.cn).}
\thanks{Siyu Chen and Fei Guo are with the School of Computer Science and Engineering, Central South University, Changsha 410083, China (e-mail: csy619@csu.edu.cn, guofei@csu.edu.cn).}
\thanks{Shiqiang Ma is with the Shenzhen Institutes of Advanced Technology, Chinese Academy of Sciences, Shenzhen 518055, China (e-mail: sq.ma@siat.ac.cn).}
\thanks{Jing Wang is with the Advanced Technology Center Beijing AI Laboratory, Chao-Yang District, Beijing 100027, China (e-mail: jingd.wang@sony.com).}
\thanks{Shengfeng He is with the School of Computing and Information Systems, Singapore Management University, Singapore 188065 (e-mail:
shengfenghe@smu.edu.sg).}}

\markboth{IEEE TRANSACTIONS ON MULTIMEDIA}%
{Shell \MakeLowercase{\textit{et al.}}: A Sample Article Using IEEEtran.cls for IEEE Journals}


\maketitle

\begin{abstract}
Large-scale Vision-Language Models like CLIP have demonstrated impressive open-set localization capabilities at the image level. However, adapting this capability to pixel-level dense prediction poses challenges due to global feature biases. In this paper, we introduce CLIPix, a simple yet effective framework that \enquote{repurposes} CLIP to perform pixel-level localization. By tracing back CLIP’s classification process, CLIPix identifies object-specific attentive regions and repurposes them as pixel-level localization cues. To address noise introduced by global biases, we propose a Noise-Resistant Correction strategy, refining these cues for more precise segmentation. Additionally, we introduce a Localization Embedding strategy to integrate both localization and enriched detail information, enabling accurate, high-resolution segmentation. Our approach preserves CLIP's generalization strength and unlocks its potential for segmenting arbitrary objects. Extensive experiments on the PASCAL and COCO datasets demonstrate that CLIPix achieves state-of-the-art performance, underscoring its effectiveness. Our code is available at \href{https://github.com/aqingaqinghh/CLIPix} {github.com/aqingaqinghh/CLIPix}. 
\end{abstract}


\begin{figure*}[t]
\centering
\includegraphics[width=\linewidth]{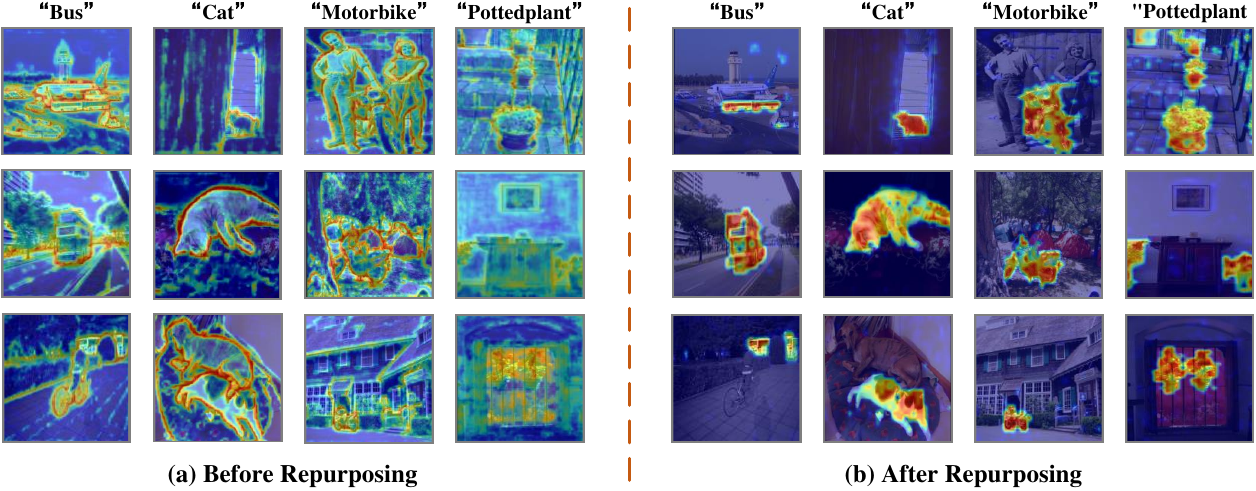}
\caption{The overview of our CLIPix framework. CLIPix repurposes CLIP for precise pixel-level localization while preserving its inherent generalization capabilities, unlocking its potential for segmentation. By tracing back to CLIP’s visual-language logits, we extract perception regions from key layers as initial localization information. Using our Noise-Resistant Correction strategy, we refine this information to reduce noise. These refined localization cues are integrated into our Localization Embedding strategy, enriching image features with detailed object perception and enhancing segmentation accuracy.}
\label{fig1}
\end{figure*}

\section{Introduction}
The rise of large-scale datasets and enhanced computational power has fueled the development of expansive pre-trained models \cite{1,98,5}, renowned for their impressive generalization abilities \cite{8}. These models have led to significant advancements in semantic segmentation, with particular attention to large visual-language models like CLIP \cite{5}. While models such as the Segment Anything Model (SAM) encounter limitations, such as manual prompt inefficiencies and false positives in automated prompting \cite{6,7,9}, CLIP offers a promising alternative with higher efficiency and generalization potential \cite{13}. By training on extensive paired image-caption datasets from the internet, CLIP effectively aligns visual and language spaces, enabling the model to locate objects based on object category names, thus addressing the inefficiencies associated with manual prompts. To enhance the applicability of CLIP models in practical edge deployments, researchers have begun to integrate Spiking Neural Networks (SNNs) \cite{106,107,108} into CLIP-based multimodal applications to improve computational efficiency. SpikeCLIP \cite{106} employs knowledge distillation for cross-modal alignment during pre-training and utilizes surrogate gradients for dual-loss fine-tuning, demonstrating that multimodal features extracted from text and images can be effectively aligned using spike trains. SPKLIP \cite{107} adopts spike-text contrastive learning to directly align raw spike streams with text and incorporates SNN principles to develop a fully spiking visual encoder, showcasing significant energy-saving advantages. SeLHIB \cite{108} extends the information bottleneck principle to self-supervised learning in SNNs, enabling robust feature learning from optical flow information in spike sequences in a flexible manner, thereby providing further insights into this field.

Recent research has focused on expanding CLIP’s classification capabilities into the realm of segmentation \cite{13,56,57}, facilitating segmentation of specific objects through simple text prompts. This paradigm aims to overcome the limitations of traditional closed-vocabulary models. Several studies \cite{56,57} have attempted to map vision and language spaces onto the same pixel grid, achieving pixel-level alignment for robust representation. However, this approach, reliant on strong class-aware alignment, risks overfitting and may struggle to recognize unseen classes. Other methods \cite{8,13} leverage the affinity matrix between vision and language to locate objects, but as illustrated in Fig.~\ref{fig1}(a), coarse localization at the pixel level can lead to false positives, thus impairing segmentation accuracy. Additional strategies \cite{16} involve employing supplementary segmentation models to generate mask proposals, using CLIP's classification capabilities for refinement. However, these methods are constrained by a reliance on closed-set segmentation models, which hampers their generalization and efficiency.

Motivated by these challenges, we pose a fundamental question: \textit{Can we repurposes CLIP to retain its strong generalization capabilities while achieving precise pixel-level localization?} In response, we introduce CLIPix, a novel framework that leverages CLIP's classification backpropagation to reveal precise regions of interest, as shown in Fig.~\ref{fig1}(b). Our method enhances CLIP’s segmentation potential by extracting class-specific activation maps from visual-language logits as a foundation for localization. Given that CLIP, originally designed for classification, emphasizes global features, pixel-level localization tends to be noisy. To address this, we propose a Noise-Resistant Correction strategy, which redistributes patch weights based on initial localization and constructs class prototypes to refine class tokens, thereby reducing noise and enhancing object specificity.

To further improve detail retention, we develop a Localization Embedding strategy that embeds localization data from CLIP into the image features, generating refined high-level features that enhance object awareness. This enriched localization information is then merged with updated image features for decoding, preserving initial localization while enhancing finer details. Our approach maintains CLIP’s original generalization strength, unlocking its potential for precise, diverse object segmentation.

Our contributions are as follows:

\begin{itemize}
    \item We introduce CLIPix, a novel framework that enables CLIP to achieve precise pixel-level localization while retaining its inherent generalization capabilities.
    \item We propose a Noise-Resistant Correction strategy that reduces noise in localization, allowing for robust and specific object segmentation.
    \item We develop a Localization Embedding strategy that enhances fine detail accuracy in segmentation by embedding localization data with image features.
    \item Our approach achieves state-of-the-art performance, surpassing existing methods on the PASCAL and COCO datasets.
\end{itemize}

\section{Related Work}
\label{sec:formatting}
\subsection{Vision-Language Pre-trained Models}
The advent of large-scale datasets and powerful computational resources has propelled the development of pre-trained vision-language models that bridge visual concepts and textual descriptions \cite{5,24}. Among these, CLIP \cite{5} stands out due to its exceptional generalization capabilities. Trained on over 400 million image-text pairs sourced from the internet, CLIP effectively links language to images, demonstrating robust image-level localization abilities and enabling diverse computer vision tasks to incorporate language understanding \cite{26, 30, 99, 100}.

Various methods \cite{73, 74} based on CLIP have propelled the development of vision-language models (VLMs). Some methods focusing on improving CLIP's multimodal fusion approach to construct more effective VLMs. CoOp \cite{71} and CoCoOp \cite{72} enhance vision-language alignment by generating diverse textual prompt templates through learnable textual prompts. Maple \cite{73} proposes a multimodal adapter to aggregate visual and linguistic features into a shared space, while simultaneously learning prompts for both image and text branches. Additionally, some methods leverage knowledge distillation techniques, using CLIP as a teacher model to fully tap into its knowledge potential. CLIP-KD \cite{74} demonstrates that the simplest feature mimicry using Mean Squared Error (MSE) loss yields the best results.

However, these efforts are all dedicated to enhancing CLIP's image-level representation capabilities. In this work, we extend CLIP’s open-set image-level localization capabilities to pixel-level localization, enabling precise segmentation of arbitrary object categories without being confined to closed-set scenarios.
\subsection{Binary Open-Set Semantic Segmentation}
Binary open-set semantic segmentation \cite{60, 9, 8, 101} focuses on segmenting any specified object category in an image, encompassing both known and unseen classes. ZS3Net \cite{60} addresses this task by combining a generative approach with pre-trained word embeddings to represent visual features of unseen objects. SPNet \cite{33} leverages knowledge similarity among known categories to transfer knowledge to unknown classes. With the rise of large-scale models, approaches like Matcher \cite{6}, Per-SAM \cite{7}, and GRP-SAM \cite{9} employ automated prompts to overcome the inefficiencies of SAM’s manual prompting. However, these methods often suffer from false positives and depend on additional reference images with specific category masks, limiting their flexibility.

More recent work has focused on leveraging CLIP \cite{5} for open-set segmentation tasks. LSeg \cite{56} introduces a language-driven segmentation model using contrastive training between pixel and category text features. SAZS \cite{57} incorporates shape awareness to enhance segmentation performance for unseen classes. PixelCLIP \cite{35} adapts CLIP’s image encoder for pixel-level understanding. However, these methods often rely on strong category-aware alignment, which can lead to overfitting. DenseCLIP \cite{13} and PGMA-Net \cite{8} utilize CLIP’s affinity matrix to provide localization cues for feature matching, but their reliance on coarse-grained prior knowledge can result in inaccurate segmentation guidance.

Our work builds upon these efforts by repurposing CLIP to achieve precise pixel-level localization while preserving its generalization capabilities. By addressing challenges related to noise and detailed localization, we refine CLIP’s utility for open-set dense prediction, unlocking its potential for accurate segmentation across arbitrary categories.

\section{method}

\begin{figure*}[t]
\centering
\includegraphics[width=1.0\linewidth]{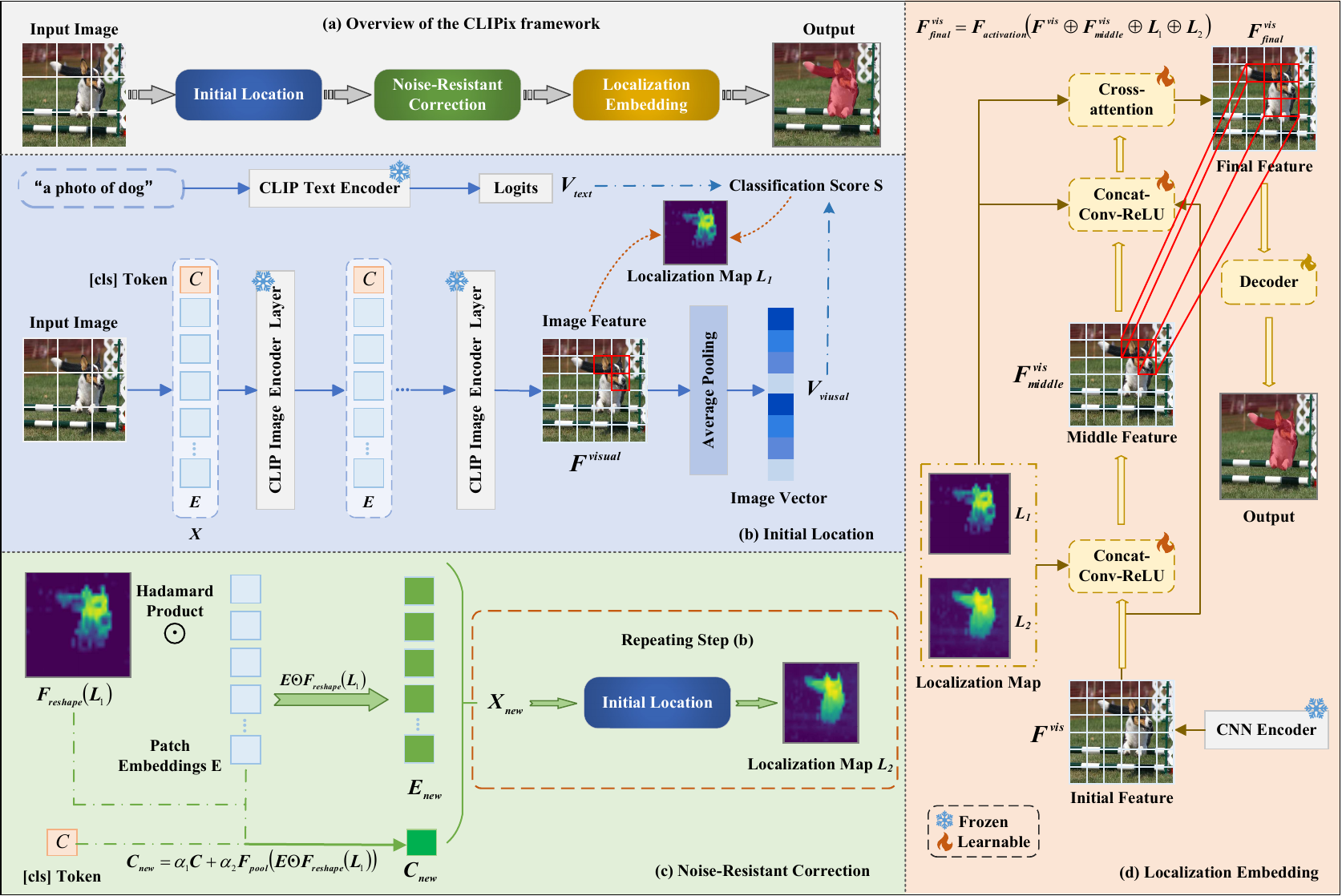}
\caption{The overview of our CLIPix framework (a). CLIPix repurposes CLIP for accurate pixel-level localization while preserving its natural generalization, unlocking its segmentation capability. From CLIP’s visual-language logits, we derive perception regions in key layers as initial localization hints (b). Our Noise-Resistant Correction strategy denoises these hints through refinement (c). The cleaned localization cues are integrated via our Localization Embedding strategy, enhancing image features with detailed object awareness and raising segmentation performance (d).}
\label{fig2}
\end{figure*}

Binary open-set semantic segmentation aims to segment objects of specific categories in an image, including those from unseen classes not encountered during training. Formally, given a dataset $D$, it is divided into a training set $D_{\text{train}} = \left\{\left(I, M, C_{\text{train}}\right)\right\}$ and a test set $D_{\text{test}}=\left\{\left(I, M, C_{\text{test}}\right)\right\}$, where $I$ and $M$ represent the input image and its corresponding ground truth mask, respectively, and $C$ denotes the target classes. Importantly, the target classes in $C_{\text{train}}$ and $C_{\text{test}}$ are strictly disjoint, ensuring a rigorous evaluation of the model's open-set capabilities.

The model is fine-tuned on $D_{\text{train}}$, after which its parameters are fixed for evaluation on $D_{\text{test}}$. This setup tests the model's ability to generalize knowledge from a limited set of training categories ($C_{\text{train}}$) to unseen categories ($C_{\text{test}}$), offering a robust assessment of its open-set segmentation performance.

\subsection{Method Overview}
Our objective is to repurpose the visual-language model CLIP to achieve more precise pixel-level localization, unlocking its potential for segmenting objects of any target class. Figure \ref{fig2} illustrates an overview of our method. First, we trace back CLIP’s classification process to extract attention regions corresponding to specific object categories. Next, we introduce a Noise-Resistant Correction strategy to refine these regions by mitigating noise caused by global biases. To further enhance localization accuracy, we propose a Localization Embedding strategy that retains the original localization cues while enriching fine details. 

\subsection{Noise-Resistant Correction}
Due to the open-set setting of binary open semantic segmentation, the model inevitably lacks perception of unseen class objects. Therefore, how to effectively extract prior localization information for perceiving unseen class objects becomes a key challenge. This challenge implies that the segmentation performance of the model largely depends on the quality of the prior localization information. For instance, in an image containing both \enquote{human} and \enquote{horse}, we aim to segment the specific class object of \enquote{human}. However, if the prior localization falls on the \enquote{horse}, it will be difficult for the model to focus on the true target class \enquote{human}.

Therefore, we need to repurpose CLIP to provide robust pixel-level prior localization information. Previous work \cite{13,8} has offered a direct method by computing the affinity matrix between text features and image feature maps to obtain pixel-level localization information. However, CLIP originates from contrastive training between global image features and textual features, which leads to its deficiencies in fine-grained visual-language alignment, making it challenging for the affinity matrix to achieve precise localization. We believe that extracting regions of interest from CLIP's mature classification decisions better represents its potential for dense localization.

Specifically, we use a simple text prompt template \enquote{a photo of $[cls]$} to convey to the CLIP text encoder to obtain the text vector $V^{text} \in R^{1\times D}$, where $D$ denotes the feature dimension, $[cls]$ represents the target class name. For a given input image $I$, CLIP initially encodes it to obtain the input $X=[E, C], X\in R^{(N+1)\times D}$, for the first VIT layer, where $E \in R^{N\times D}$ representing the image features, and $C \in R^{1\times D}$ token denotes the class token of the visual encoder. Subsequently, the initial input $X$ is passed to the subsequent CLIP visual encoder to obtain the final image features $F^{visual} \in R^{HW\times D}$, where HW equals N. Then, $F^{visual}$ undergoes average pooling to derive the image vector $V^{visual}$, which is used to compute the classification score $S$ with $V^{text} \in R^{1\times D}$.

\begin{equation}
S=\frac{V_{text}^{T} V_{visual}}{\left\|V_{text}\right\|\left\|V_{visual}\right\|}
\end{equation}
where $T$ denotes the transpose of a matrix. Next, we trace back the response activation maps $L_{1} \in R^{H\times W}$ of the key layers in the process of CLIP computing $S$ to provide prior localization information.

\begin{equation}
L_{1}=\operatorname{ReLU}\left(\sum_{m} F_{visual}^{m} \frac{1}{HW} \sum_{i} \sum_{j} \frac{\partial S}{\partial F_{visual}^{m}(i, j)}\right)
\end{equation}
where $F_{visual}^{m} \in R^{H\times W}$ represents the activation value of the $m-th$ feature map. Compared to the ambiguous localization caused by forced alignment in the affinity matrix, the response activation map $L_{1}$ can provide more accurate localization. However, since CLIP prefers to align global features with textual features and is insensitive to local noise, it may respond to some non-target areas. To address this issue, we propose a high-response correction strategy. Based on the initial localization information $L_{1}$, we re-adjust the initial input $X$ to enhance the response of the target area during the feature extraction stage and suppress local noise. Specifically, $L_{1}$ is used as weight information to adjust the distribution of the image input features $E$.

\begin{equation}
E_{new}=E \odot \mathcal{F}_{\text{reshape}}\left(L_{1}\right)
\end{equation}
where $\odot$ represents Hadamard product, and $\mathcal{F}_{\text{reshape}}$ reshapes the size of the input tensor to $N\times D$. Furthermore, we integrate $L_{1}$ and $E$ to construct a pseudo-target prototype for updating the class token C, further enhancing the perception of the target class.

\begin{equation}
C_{new}=\alpha_{1}C+\alpha_{2}\mathcal{F}_{\text{pool}}\left(E \odot \mathcal{F}_{\text{reshape}}\left(L_{1}\right)\right)
\label{4}
\end{equation}
where $\alpha_{1}$ and $\alpha_{2}$ are set to 0.5 and 0.5, respectively, $\mathcal{F}_{\text{pool}}$ represents the average-pooling operation. At this point, the initial input $X$ has been updated to a new input $X_{new}=\left[E_{new}, C_{new}\right]$ that perceives the target. Subsequently, $X_{new}$ is passed through the visual encoder of CLIP, and by repeating the aforementioned steps, corrected pixel-level localization information $L_{2} \in R^{H\times W}$ with high response to the target object can be obtained.

\subsection{Localization Embedding }

CLIP possesses sophisticated classification capabilities, and by tracing the response regions of key layers to target class objects during the classification process, high-quality critical localizations can be obtained. However, we observe that it tends to respond to key local features of the target class objects to complete category judgment, such as the \enquote{head} region of a \enquote{horse} rather than the whole body. Although this preference can achieve success in classification tasks, it may introduce local biases in dense prediction tasks, leading to incomplete segmentation.

Specifically, due to the open-set setting, the model has not undergone tuning for unseen categories, resulting in inadequate perception of unseen category objects and a lack of knowledge about the relationships among target object components. While prior localization information provides localization of partial regions of the target object, the model is unable to perceive other regions of the target object.

To address this limitation, we have designed a Localization Embedding strategy that not only retains the original localization information but also enhances localization details for precise full-body prediction of the target object. Our approach is inspired by the idea of skip connections, employing cascaded activation and aggregation decoding strategies. Specifically, due to the weak perception of the initial feature map $F^{vis}$ regarding unknown class objects, we need to globally activate it. Therefore, we construct localization prototypes $P_{1} \in R^{1\times D}$ and $P_{2} \in R^{1\times D}$ based on localization information $L_{1}$ and $L_{2}$, respectively, to condense the key information of the target object.
\begin{equation}
P_{1}, P_{2} = \mathcal{F}_{\text{pool}}\left(F^{vis} \odot L_{1}\right), \mathcal{F}_{\text{pool}}\left(F^{vis} \odot L_{2}\right)
\end{equation}

Then, we expand $P_{1}$ and $P_{2}$, concatenate them with the initial image feature map $F^{vis}$, and activate them with a larger receptive field to obtain an intermediate image feature map $F_{middle}^{vis} \in R^{HW\times D}$.

\begin{equation}
F_{middle}^{vis} = \mathcal{F}_{\text{activation}}\left(F^{vis} \oplus \mathcal{F}_{\text{repeat}}\left(P_{1}\right)  \oplus \mathcal{F}_{\text{repeat}}\left(P_{2}\right) \right)
\end{equation}
where $\oplus$ is the concatenation operation in channel dimension, $\mathcal{F}_{\text{activation}}$ is composed solely of a convolution and ReLU, and $\mathcal{F}_{\text{repeat}}$ repeat the size of the input tensor to $HW\times D$. At this point, the intermediate image feature map $F_{middle}^{vis}$ has initially acquired the ability to perceive the relationships among the components of the target object. Furthermore, we utilize the key localizations $L_{1}$ and $L_{2}$ to finally activate $F_{middle}^{vis}$, obtaining the final image feature map $F_{final}^{vis} \in R^{HW\times D}$. This allows us to expand from local perception to global perception. Meanwhile, during the activation process, we still use the initial image feature map $F^{vis}$ to prevent information forgetting.

\begin{equation}
F_{final}^{vis} = \mathcal{F}_{\text{activation}}\left(F^{vis} \oplus F_{middle}^{vis}  \oplus L_{1} \oplus L_{2} \right)
\end{equation}

At this point, we have gathered abundant elements that can be provided to the decoding operation. Finally, we have designed a final prediction module based on the transformer decoder. Specifically, we utilize localization information $L_{2}$ and the intermediate image feature map $F_{middle}^{vis}$ to respond to key local features, serving as representatives of class-specific features. These are then used to compute a cross-attention correlation map $M \in R^{N\times N}$ with the final image feature map $F_{final}^{vis}$, enabling the extension of local localization to full-body localization.

\begin{equation}
M=\operatorname{softmax} \left(\frac{\left\langle W^{q} F_{final}^{vis}, W^{k}\left(F_{middle}^{vis} \odot L_{2}\right)\right\rangle}{\left\|W^{q} F_{final}^{v is}\right\|\left\|W^{k}\left(F_{middle}^{vis} \odot L_{2}\right)\right\|}\right)
\end{equation}
where $W^{q} \in R^{D\times D}$, $W^{k}\in R^{D\times D}$ denote the learnable parameters, $\langle \rangle$ represents the calculation of cosine similarity. Next, we integrate the attention score maps $M$ and concatenate them with the prior localization information $P_{1}$ and $P_{2}$ to achieve comprehensive full-body target localization.

\begin{equation}
F^{out}=W^{o}\left(M\left(W^{v}\left(F_{middle}^{vis} \odot L_{2}\right)\right) \oplus L_{1} \oplus L_{2}\right)
\end{equation}
where $W^{v} \in R^{D\times D}$, $W^{o} \in R^{D\times (D+2)}$ denote the learnable parameters. Finally, we pass the activated output features $F^{out} \in R^{HW\times D}$ to a classifier for decoding, in order to make the final prediction. Owing to the high target-aware capability already embedded in the image features at this stage, our decoder comprises only a few convolution and normalization operations to achieve efficient segmentation. We use this final prediction along with the ground truth labels to construct a cross-entropy loss function for training the model.

\section{Experiments}
\begin{table*}[t]
\centering 
\caption{Performance comparison on PASCAL-5$^{i}$ and COCO-20$^{i}$ in terms of mIoU(\%). Results in \textbf{bold} denote the best performance on zero-shot setting. * indicates that this method utilizes a multimodal model.}
\label{tab1}
\resizebox{2.0\columnwidth}{!}
{\footnotesize
\begin{tabular}{l|c|c|ccccc|ccccc}
\hline
\multirow{2}{*}{\textbf{Methods}} & \multirow{2}{*}{\textbf{Backbone}} & \multirow{2}{*}{\textbf{Setting}} & \multicolumn{5}{c|}{\textbf{PASCAL-5$^{i}$}} & \multicolumn{5}{c}{\textbf{COCO-20$^{i}$}}\\
& & &5$^{0}$&5$^{1}$&5$^{2}$&5$^{3}$&\textbf{Mean}&20$^{0}$&20$^{1}$&20$^{2}$&20$^{3}$&\textbf{Mean}\\ 
\hline
\hline
HSNet (ICCV'21) \cite{23} & ResNet &\multirow{9}{*}{1-shot} &67.3 &72.3 &62.0 &63.1 &66.2 &37.2 &44.1 &42.4 &41.3 &41.2 \\
SSP (ECCV'22) \cite{19} & ResNet &\multirow{8}{*}{} &63.2 &70.4 &68.5 &56.3 &64.6 &39.1 &45.1 &42.7 &41.2 &42.0 \\
BAM (CVPR'22) \cite{20} & ResNet &\multirow{8}{*}{} &68.9 &73.6 &67.5 &61.3 &67.8 &43.4 &50.6 &47.5 &43.4 &46.2 \\
MIANet (CVPR'23) \cite{21} & ResNet &\multirow{8}{*}{} &68.5 &75.7 &67.5 &64.2 &68.7 &42.5 &52.9 &47.7 &47.7 &47.6 \\
HDMNet (CVPR'23) \cite{61} & ResNet &\multirow{8}{*}{} &71.0 &75.4 &68.9 &62.1 &69.4 &43.8 &55.3 &51.6 &49.4 &50.0 \\
HMNet (NIPS'24)\cite{67} & ResNet &\multirow{8}{*}{} &72.2 &75.4 &70.0 &63.9 &70.4 &45.5 &58.7 &52.9 &51.4 &52.1 \\
AENet (ECCV'24) \cite{68} & ResNet &\multirow{8}{*}{} &71.3 &75.9 &68.6 &65.4 &70.3 &45.4 &57.1 &52.6 &50.0 &51.3 \\
ABCB (CVPR'24) \cite{62} & ResNet &\multirow{8}{*}{} &73.0 &76.0 &69.7 &69.2 &72.0 &46.0 &56.3 &54.3 &51.3 &51.5 \\
PI$\_$CLIP* (CVPR'24) \cite{63} & ResNet &\multirow{9}{*}{} &76.4 &83.5 &74.7 &72.8 &76.8 &49.3 &65.7 &55.8 &56.3 &56.8 \\
LLaFS++* (TPAMI'25) \cite{110} & ResNet &\multirow{9}{*}{} &77.8 &82.1 &75.8 &72.9 &77.2 &50.8 &62.7 &60.2 &56.4 &57.5 \\
DSV-LFS* (CVPR'25) \cite{109} & SAM &\multirow{1}{*}{} &71.6 & 81.9 & 71.1 & 75.0 & 74.9 & 69.9 & 73.3 & 70.6 & 71.3 & 71.3 \\
\hline
\hline
HSNet (ICCV'21) \cite{23} &ResNet &\multirow{9}{*}{5-shot} &71.8 &74.4 &67.0 &68.3 &70.4 &45.9 &53.0 &51.8 &47.1 &49.5 \\
SSP (ECCV'22) \cite{19} &ResNet &\multirow{8}{*}{} &70.5 &76.4 &79.0 &66.4 &73.1 &47.4 &54.5 &50.4 &49.6 &50.2 \\
BAM (CVPR'22) \cite{20} &ResNet &\multirow{8}{*}{} &70.6 &75.0 &70.8 &67.2 &70.9 &49.2 &54.2 &51.6 &49.5 &51.1 \\
MIANet (CVPR'23) \cite{21} &ResNet &\multirow{8}{*}{} &70.2 &77.4 &70.0 &68.7 &71.6 &45.8 &58.1 &51.2 &51.9 &51.6 \\
HDMNet (CVPR'23) \cite{61} &ResNet &\multirow{8}{*}{} &71.3 &76.2 &71.3 &68.5 &71.8 &50.6 &61.6 &55.7 &56.0 &56.0 \\
HMNet (NIPS'24) \cite{67} &ResNet &\multirow{8}{*}{} &74.2 &77.3 &74.1 &70.9 &74.1 &53.4 &64.6 &60.8 &56.8 &58.9 \\
AENet (ECCV'24) \cite{68} &ResNet &\multirow{8}{*}{} &73.9 &77.8 &73.3 &72.0 &74.2 &52.7 &62.6 &56.8 &56.1 &57.1 \\
ABCB (CVPR'24) \cite{62} &ResNet &\multirow{8}{*}{} &74.8 &78.5 &73.6 &72.6 &74.9 &51.6 &63.5 &62.8 &57.2 &58.8 \\
PI$\_$CLIP* (CVPR'24) \cite{63} &ResNet &\multirow{8}{*}{} &76.7 &83.8 &75.2 &73.2 &77.2 &56.4 &66.2 &55.9 &58.0 &59.1 \\
LLaFS++* (TPAMI'25) \cite{110} &ResNet &\multirow{9}{*}{} &79.7 &83.6 &77.9 &73.8 &78.8 &53.9 &64.9 &63.8 &61.1 &60.9 \\
DSV-LFS* (CVPR'25) \cite{109} & SAM &\multirow{1}{*}{} &72.0 & 82.0 & 71.3 & 75.5 & 75.2 & 71.0 & 73.8 & 71.3 & 71.4 & 71.9 \\
\hline
\hline
ZS3Net* (NeurIPS'19) \cite{60} &{ResNet} &\multirow{16}{*}{0-shot} &40.8 &39.4 &39.3 &33.6 &38.3 &18.8 &20.1 &24.8 &20.5 &21.1 \\
LSeg* (ICLR'22) \cite{56} &{ResNet} &\multirow{8}{*}{} &52.8 &53.8 &44.4 &38.5 &47.4 &22.1 &25.1 &24.9 &21.6 &23.4 \\
PFENet* (TPAMI'22) \cite{59} &{ResNet} &\multirow{8}{*}{} &50.0 &68.5 &51.7 &46.6 &54.2 &- &- &- &- &- \\
HPA* (TPAMI'22) \cite{82} &{ResNet} &\multirow{8}{*}{} &52.7 &70.1 &60.4 &51.7 &58.7 &- &- &- &- &- \\
BAM* (TPAMI'23) \cite{58} &{ResNet} &\multirow{8}{*}{} &52.0 &70.2 &59.4 &49.0 &57.7 &- &- &- &- &- \\
SAZS* (CVPR'23) \cite{57} &{DRN} &\multirow{8}{*}{} &57.3 &60.3 &58.4 &45.9 &55.5 &34.2 &36.5 &34.6 &35.6 &35.2 \\
CLIPSeg* (CVPR'22) \cite{81} &{ViT-B} &\multirow{9}{*}{} &53.9 &62.0 &42.8 &48.0 &51.6 &34.2 &38.9 &34.9 &31.9 &34.9 \\
LSeg* (ICLR'22) \cite{56} &{ViT-L} &\multirow{8}{*}{} &61.3 &63.6 &43.1 &41.0 &52.3 &28.1 &27.5 &30.0 &23.2 &27.2 \\
UniBoost* (arXiv'2023) \cite{80} &{ViT-L} &\multirow{9}{*}{} &67.3 &65.1 &46.7 &47.3 &56.6 &30.4 &31.9 &35.7 &33.5 &32.8 \\
SAZS* (CVPR'23) \cite{57} &{ViT-L} &\multirow{8}{*}{} &62.7 &64.3 &60.6 &50.2 &59.4 &33.8 &38.1 &34.4 &35.0 &35.3 \\
PAT* (TPAMI'24) \cite{79} &{ViT-B} &\multirow{8}{*}{} &67.0 &69.6 &56.4 &51.8 &61.2 &28.4 &36.2 &32.5 &33.0 &32.5 \\
PAT* (TPAMI'24) \cite{79} &{DeiT-B} &\multirow{8}{*}{} &70.1 &70.6 &54.1 &54.3 &62.3 &30.1 &37.9 &37.5 &34.3 &34.9 \\
PMGA-Net* (TMM'24) \cite{8} &{ResNet} &\multirow{8}{*}{} &68.2 &78.8 &68.8 &66.5 &70.6 &- &- &- &- &- \\
\rowcolor{mygray}\textbf{Ours*} &{ResNet} & {} &\textbf{77.9} &\textbf{87.5} & \textbf{79.7} &\textbf{77.5} &\textbf{80.7} &\textbf{58.8} &\textbf{64.9} &\textbf{57.8} &\textbf{65.7} &\textbf{61.8} \\
\rowcolor{mygray}\textbf{Ours*} &{MobileNet} & {} &75.1 &83.7 & 75.8 &73.1 &76.9 &54.9 &60.7 &53.8 &60.4 &57.5 \\
\rowcolor{mygray}\textbf{Ours*} &{EfficientNet} & {} &76.9 &85.9 & 77.9 &75.8 &79.1 &56.4 &62.1 &55.1 &62.9 &59.1 \\
\hline
\end{tabular}
 }
\end{table*}

\begin{table}[t]
\centering 
\caption{Performance comparison with state-of-the-art on PASCAL-5$^{i}$ and COCO-20$^{i}$ in terms of FBIoU(\%). Results in bold denote the best performance.}
\label{tab2}
\resizebox{1.0\columnwidth}{!}{
\begin{tabular}{l|c|cc}
\hline
\multirow{2}{*}{\textbf{Methods}} & \multirow{2}{*}{\textbf{Setting}} & \multicolumn{2}{c}{\textbf{FBIoU}}\\ & &PASCAL-5$^{i}$&COCO-20$^{i}$\\ 
\hline
\hline
HSNet (ICCV'21) \cite{23} &\multirow{6}{*}{5-shot} &80.6 &72.4 \\
DACM (ECCV'22) \cite{69} &\multirow{8}{*}{} &81.5 &71.6 \\
MIANet (CVPR'23) \cite{21} &\multirow{8}{*}{} &82.2 &73.1 \\
HDMNet (CVPR'23) \cite{61} &\multirow{8}{*}{} &- &77.7 \\
HMNet (NIPS'24)\cite{67} &\multirow{8}{*}{} &84.4 &77.6 \\
AENet (ECCV'24) \cite{68} &\multirow{8}{*}{} &84.5 &78.5 \\
\hline
\hline
HSNet (ICCV'21) \cite{23} &\multirow{6}{*}{1-shot} &77.6 &69.1 \\
DACM (ECCV'22) \cite{69} &\multirow{8}{*}{} &78.9 &68.9 \\
MIANet (CVPR'23) \cite{21} &\multirow{8}{*}{} &79.5 &71.5 \\
HDMNet (CVPR'23) \cite{61} &\multirow{8}{*}{} &- &72.2 \\
HMNet (NIPS'24) \cite{67} &\multirow{8}{*}{} &81.6 &74.5 \\
AENet (ECCV'24) \cite{68} &\multirow{8}{*}{} &81.2 &74.4 \\
\hline
\hline
SPNet (CVPR'19) \cite{70} &\multirow{7}{*}{0-shot} &44.3 &- \\
ZS3Net (NeurIPS'19) \cite{60} &\multirow{7}{*} &57.7 &55.1 \\
LSeg (ICLR'22) \cite{56} &\multirow{8}{*}{} &67.6 &59.9 \\
SAZS (CVPR'23) \cite{57} &\multirow{8}{*}{} &69.0 &58.2 \\
PMGA-Net (TMM'24) \cite{8} &\multirow{8}{*}{} &80.0 &- \\
PAT (TPAMI'24) \cite{79} &\multirow{8}{*}{} &75.1 & 57.7 \\
\rowcolor{mygray}\textbf{CLIPix(ours)} &{} &\textbf{88.4} &\textbf{78.8} \\
\hline
\end{tabular}
 }
\end{table}

\begin{figure*}[t]
\centering
\includegraphics[width=0.9\linewidth]{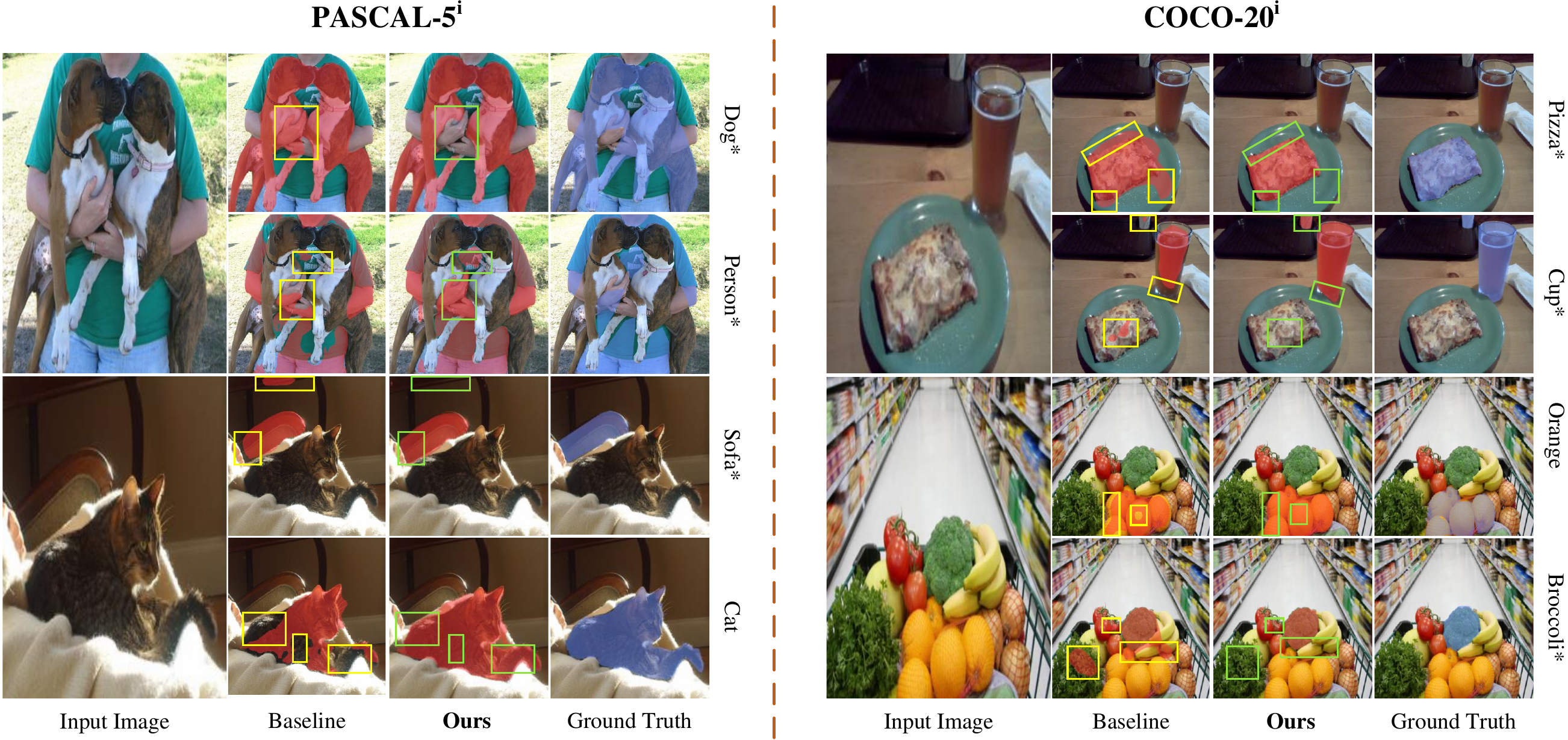}
\caption{Qualitative results of the proposed method and baseline approach on PASCAL-5$^{i}$ and COCO-20$^{i}$. From left to right: input image, prediction of baseline, prediction of our method CLIPix, ground truth. * denotes unseen classes during training.}
\label{fig3}
\end{figure*}

\subsection{Dataset and Evaluation Metrics}
To validate the segmentation performance and generalization capability of our CLIPix. We conducted extensive experiments on two widely used datasets for binary open-set semantic segmentation tasks, namely PASCAL-5$^{i}$ \cite{36} and COCO-20$^{i}$ \cite{37}. Specifically,  PASCAL-5$^{i}$ consists of PASCAL VOC 2012 and is expanded by SBD, with a total of 20 classes and divided into four folds: 5$^{0}$, 5$^{1}$, 5$^{2}$, and 5$^{3}$. COCO-20$^{i}$ is a more challenging dataset based on MS COCO, with a total of $80$ classes and divided into four folds: 20$^{0}$, 20$^{1}$, 20$^{2}$, and 20$^{3}$. In the experiments, one fold is used for evaluation, while the others are used for training. To ensure a fair comparison, like the evaluation protocol widely used in previous works, we use average intersection over union (mIoU) as evaluation metrics.

\subsection{Implementation Details}
We use the pre-trained CLIP-ViT-B/16 \cite{1} as our backbone. We additionally constructed a pre-trained ResNet50 (or MobileNetV2 \cite{104} / EfficientNet-B4 \cite{105}) to serve as the feature map for image perception and decoding in the LE strategy. During training, the network is optimized by an AdamW optimizer with a momentum of 0.9 and a learning rate of
0.0001. Our models on PASCAL-5$^{i}$ and COCO-20$^{i}$ are trained for 200 epochs and 50 epochs respectively, with the batch size set to 16. All models are trained on one NVIDIA Tesla V100 GPU.

\subsection{Comparison with State-of-the-Arts}
\noindent{\bf Quantitative Results.} To evaluate the effectiveness of our method, as shown in Tables \ref{tab1} and \ref{tab2}, we compared it with other advanced methods based on zero-shot and few-shot settings in the context of binary open-set semantic segmentation tasks. We reported the results using different folds as the test set and their average outcomes. The compared methods are representative works published in the past two years. Our method essentially operates under a zero-shot setting. As Tables \ref{tab1} and \ref{tab2} indicates, our performance significantly surpasses that of other methods also based on a zero-shot setting. Among them, SAZS \cite{57} and LSeg \cite{56} align text and image features through contrastive training between pixels and class texts. However, this strongly class-aware approach risks overfitting. Our approach achieves performance upper bounds that surpass those of these works by {\bf 21.3\%} and {\bf 26.5\%} in terms of mIoU on PASCAL-5i and COCO-20i respectively, as well as by {\bf 19.4\%} and {\bf 18.9\%} in terms of FBIoU on the same datasets. Additionally, PMGA-Net \cite{8} utilizes the affinity matrix between CLIP visual and language features as a prior localization to guide segmentation predictions. Nevertheless, this coarse localization may suffer from false positives, posing a risk of misleading guidance. In contrast, the prior localization extracted by our method is robust and accurate, better guiding segmentation predictions. The advantage of {\bf 10.1\%} on PASCAL-5$^{i}$ validates this conclusion. Notably, when lightweight backbone networks such as MobileNetV2 \cite{104} and EfficientNet-B4 \cite{105} are employed, our method still significantly outperforms existing zero‑shot semantic segmentation approaches while maintaining efficient inference, demonstrating its promising potential for edge deployment.

Furthermore, we have also compared our method with approaches based on few-shot settings (including 1-shot and 5-shot). As shown in Tables \ref{tab1} and \ref{tab2}, even without utilizing any additional support information, our method still demonstrates comprehensive performance advantages, fully validating its effectiveness and practical application potential. Among these methods, DSV-LFS \cite{109} and LLaFS++ \cite{110} both leverage the latest large-scale pre-trained vision-language models and language models to acquire richer prior knowledge. Specifically, DSV-LFS utilizes GPT‑4 to generate rich and detailed textual descriptions, which are then fed into the large-scale vision-language model LLaVA‑1.5 \cite{111} to obtain category-relevant prompts. Concurrently, the method employs the vision foundation model SAM \cite{1} to extract visual features and constructs visual prompts in combination with support samples. Finally, segmentation is accomplished by inputting both the prompts and visual features into the SAM decoder. Similarly, LLaFS++ relies on the BLIPv2 \cite{112} vision-language model for dense prediction and utilizes the large language model to generate more comprehensive and accurate textual descriptions, thereby enhancing vision-language localization capabilities.

However, these methods still exhibit certain limitations. To accomplish few-shot semantic segmentation, DSV‑LFS integrates multiple large models, including GPT‑4, LLaVA‑1.5, and SAM, leading to significantly increased computational overhead; LLaFS++ faces a similar computational burden. Such approaches typically rely on large-scale computational resources, placing them at a clear disadvantage in edge deployment scenarios. Moreover, they remain affected by the global bias of vision-language models and must depend on support information to provide localization priors, which limits their applicability in open-set settings.

In contrast, our method does not rely on rich textual descriptions, thereby avoiding the additional computational costs associated with external models such as GPT‑4. We transform the direct localization problem into a retrospective mechanism based on attention information from key layers during the classification process, thereby achieving more precise localization and effectively mitigating the global bias issue in vision-language models. Leveraging these reliable localization cues, our method can be directly applied to open-set scenarios without requiring any support samples. Even under conditions of limited prior information, the performance of our method still surpasses that of the aforementioned comparative approaches. This series of advantages significantly reduces the computational burden, making it more suitable for resource-constrained edge deployment environments. It is worth noting that our method does not employ the SAM model, as this foundation model—pre-trained on a vast number of categories (including the test set of our benchmark dataset)—has evaluation results that may not fully reflect true performance in open-set settings; in contrast, our experimental setup offers greater fairness and practical evaluation value.

\noindent{\bf Qualitative Results.} Our method CLIPix repurposes CLIP to provide more accurate pixel-level localization while maintaining its inherent generalization ability, enabling the segmentation of specified class objects within images. As shown in Figures \ref{fig3}, we achieve precise segmentation of any specified class object in images on both datasets, even for unseen classes. In contrast, we constructed a baseline method using an affinity matrix as prior localization, which is limited by coarse prior localization and suffers from issues such as false positives and insufficient segmentation.

\subsection{Comparison with Foundation Models}
\noindent{\bf Quantitative Results.} To further validate the advantages of our model in binary open-set semantic segmentation, we compared it with large-scale pre-trained general segmentation models on the highly challenging COCO-20$^{i}$ dataset, as shown in Table \ref{tab3}. Among them, Painter \cite{10} and SegGPT \cite{11} were trained using all classes data from COCO-20$^{i}$, and thus had no unseen classes. However, under such unfavorable conditions, our model still outperformed the best-performing SegGPT by a margin of {\bf 5.7\%}. Additionally, PerSAM \cite{7}, Matcher \cite{6}, and VRP-SAM \cite{9} customized segmentation for SAM \cite{1} to alleviate the efficiency issue of manual prompts. Yet, leveraging the precise localization provided by the taught CLIP, our model surpassed these bulky general segmentation models by a significant margin of {\bf 7.9\%}.

It is noteworthy that these methods still require the assistance of support information. This means that when segmenting a specific object, another image containing objects of that class along with its mask still needs to be provided. Although this alleviates the efficiency issue of manual prompts, there remains inconvenience. In contrast, our method does not need to worry about this issue at all, as we only need to know the target class name to achieve segmentation prediction. This significantly demonstrates our application potential.

\begin{figure}[t]
\centering
\includegraphics[width=0.9\linewidth]{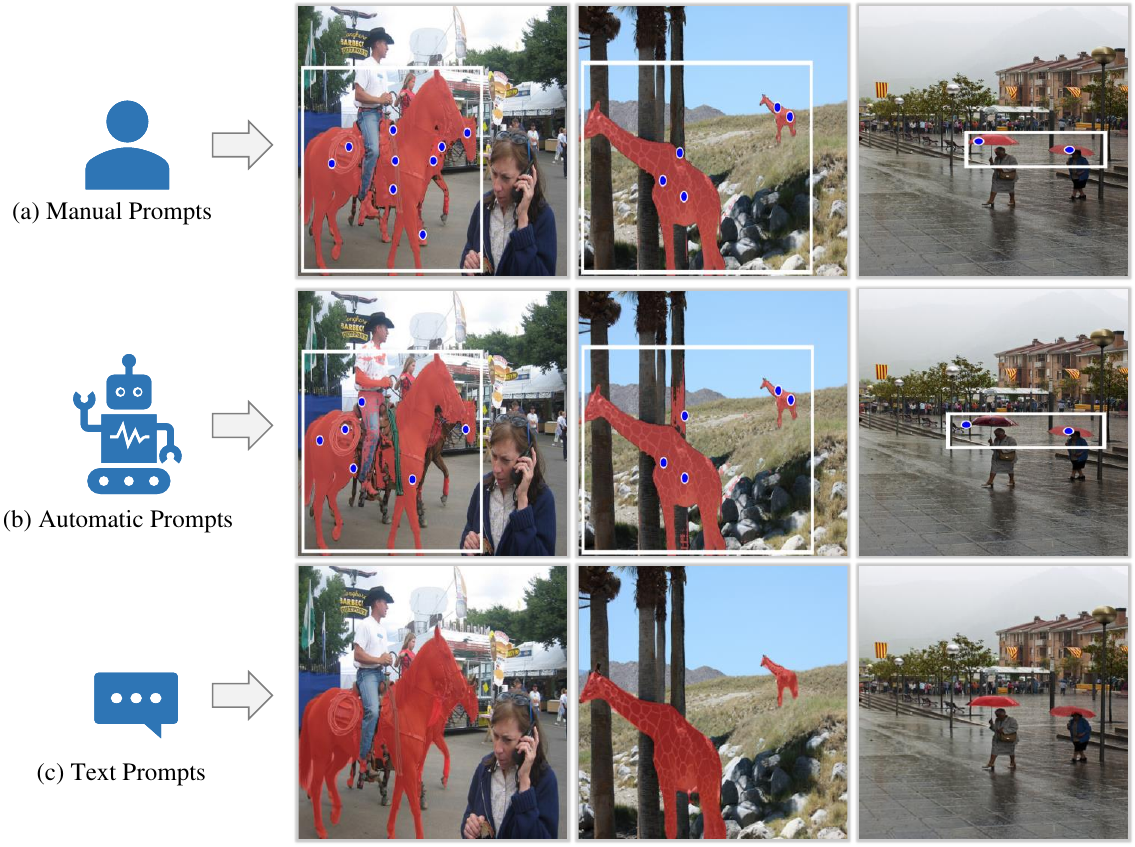}
\caption{Comparison of foundation model application modes. (a) Segmentation relies on user-provided manual cues. (b) Segmentation uses model-generated geometric cues. (c) Our method achieves segmentation using only the target class name.}
\label{fig4}
\end{figure}

\begin{table}[t]
\renewcommand\arraystretch{1.2}
\setlength{\tabcolsep}{5pt} 
\centering
\caption{Compare with other foundation models on COCO-20$^{i}$.}
\label{tab3}
\resizebox{1.0\linewidth}{!}{
    \begin{threeparttable}{}
    \begin{tabular}{c|c|ccccc}    
        \hline
        \textbf{Method} & \textbf{Prompt} & 20$^{0}$&20$^{1}$&20$^{2}$&20$^{3}$&\textbf{Mean} \\
        \hline
        Painter (CVPR'23) \cite{10} & \multirow{8}{*}{$mask$} & 31.2 & 35.3 & 33.5 & 32.4 & 33.1 \\
        SegGPT (ICCV'23) \cite{11} &   &  56.3 & 57.4 & 58.9 & 51.7 &  56.1 \\
        PerSAM (arXiv'23) \cite{7} &   &  23.1 & 23.6 & 22.0 & 23.4 & 23.0 \\
        PerSAM-F (arXiv'23) \cite{7} &   & 22.3 & 24.0 & 23.4 & 24.1 & 23.5 \\
        Matcher (ICLR'24) \cite{6} &   & 52.7 & 53.5 & 52.6 & 52.1 & 52.7 \\
        VRP-SAM (CVPR'24) \cite{9} &   & 48.1 & 55.8 & \textbf{60.0} & 51.6 & 53.9 \\
        LLaFS (CVPR'24) \cite{12} &   & 47.5 & 58.8 & 56.2 & 53.0 & 53.9 \\
        FCP (AAAI'25) \cite{84} &   & 46.4 & 56.4 & 55.3 & 51.8 & 52.5 \\
        \hline
        \rowcolor{mygray}\textbf{Ours} & $text$ &\textbf{58.8} &\textbf{64.9} &57.8 &\textbf{65.7} &\textbf{61.8} \\
        \hline
    \end{tabular}
    \end{threeparttable}}
\end{table}

\noindent{\bf Qualitative Results.} To further intuitively demonstrate the effectiveness of our method, we provide a visual comparison in Figure \ref{fig4}. It can be observed that even with manual cues, there may be issues of insufficient segmentation. Furthermore, automatic cues can suffer from false positives, and the foundation model has poor tolerance for false positive cues, which will significantly affect the segmentation quality. Compared to these methods, our method only requires the provision of the target class name to achieve robust segmentation prediction.

\begin{table}[t]
\centering
\caption{Ablation study for each component of our approach on the PASCAL-5$^{i}$ and COCO-20$^{i}$.}
\resizebox{.8\columnwidth}{!}{
    \begin{tabular}{cc|ccc}   
        \hline
        \multirow{2}{*}{\textbf{NRC}} & \multirow{2}{*}{\textbf{LE}} & \multicolumn{3}{c}{\textbf{PASCAL-5$^{i}$}} \\ & & mIoU & FB-IoU & Boundary-IoU \\
        \hline
         & & 63.0 & 79.3 & 50.3 \\
        \checkmark & & 73.0 & 82.3 & 61.7 \\
        \checkmark & \checkmark & \textbf{80.7} & \textbf{88.4} & \textbf{70.1} \\
        \hline
        \hline
        \multirow{2}{*}{\textbf{NRC}} & \multirow{2}{*}{\textbf{LE}} & \multicolumn{3}{c}{\textbf{COCO-20$^{i}$}} \\ & & mIoU & FB-IoU & Boundary-IoU \\
        \hline
         & & 40.0 & 68.8 & 36.1 \\
        \checkmark & & 53.0 & 73.8 & 45.7 \\
        \checkmark & \checkmark & \textbf{61.8} & \textbf{78.8} & \textbf{54.1} \\
        \hline
    \end{tabular}
}
\label{tab4}
\end{table}

\begin{table}[t]
\centering
\caption{Ablation study of NRC module components on the PASCAL-5$^{i}$ and COCO-20$^{i}$. $E$ and $C$ are the key elements that generate L1. $E_{new}$ and $C_{new}$ are the key elements that generate L2.}
\resizebox{.9\columnwidth}{!}{
    \begin{tabular}{cccc|cc}   
        \hline
        \multicolumn{2}{c}{\textbf{$L_{1}$}} & \multicolumn{2}{c}{\textbf{$L_{2}$}} & \multicolumn{1}{c}{\multirow{2}{*}{\textbf{PASCAL-5$^{i}$}}} & \multicolumn{1}{c}{\multirow{2}{*}{\textbf{COCO-20$^{i}$}}} \\ $E$ & $C$ & $E_{new}$ & $C_{new}$ & \multicolumn{1}{c}{} & \multicolumn{1}{c}{} \\
        \hline
        \checkmark & \checkmark &  &  & 68.0 & 48.5 \\
        \checkmark & \checkmark & \checkmark &  & 71.4 & 51.6 \\
        \checkmark & \checkmark &  & \checkmark & 70.1 & 50.1 \\
        \checkmark & \checkmark & \checkmark & \checkmark & \textbf{73.0} & \textbf{53.0} \\
        \hline
    \end{tabular}
}
\label{tab5}
\end{table}

\subsection{Ablation Study}
We conducted a series of ablation studies on the PASCAL-5$^{i}$ and COCO-20$^{i}$ datasets to investigate the effectiveness of CLIPix. We established a baseline by removing our two strategies and adopting the DenseCLIP \cite{13} paradigm, utilizing the affinity matrix as prior localization information. Subsequently, detailed ablation studies were performed based on whether the Noise-Resistant Correction strategy (NRC) and the Localization Embedding strategy (LE) were utilized.

\noindent{\bf Effects of Noise-Resistant Correction strategy.} Extracting regions of interest from the mature classification decisions of CLIP as initial information can only localize partial key information, and due to global biases, there exists the issue of localization noise. The Noise-Resistant Correction strategy (NRC) aims to address this issue. By combining the initial localization information, noise-resistant correction is achieved to obtain more robust prior localization. These two localization pieces of information work together to pave the way for subsequent feature matching, exerting a localization effect from local to global. As shown in Table \ref{tab4}, after introducing the NRC strategy, the model achieved significant improvements of {\bf 10.0\%}, {\bf 3.0\%}, and {\bf 11.4\%} in mIoU, FB‑IoU, and Boundary‑IoU on the PASCAL‑5$^{i}$ dataset, respectively. Meanwhile, on the COCO‑20$^{i}$ dataset, mIoU, FB‑IoU, and Boundary‑IoU also showed remarkable gains of {\bf 12.0\%}, {\bf 5.0\%}, and {\bf 9.6\%}, respectively. These results fully demonstrate that the module effectively enhances the quality of prior localization—not only improving the holistic perception of targets but also refining the capture of edge details, which is crucial in open‑set scenarios.

\begin{figure}[t]
\centering
\includegraphics[width=1.0\linewidth]{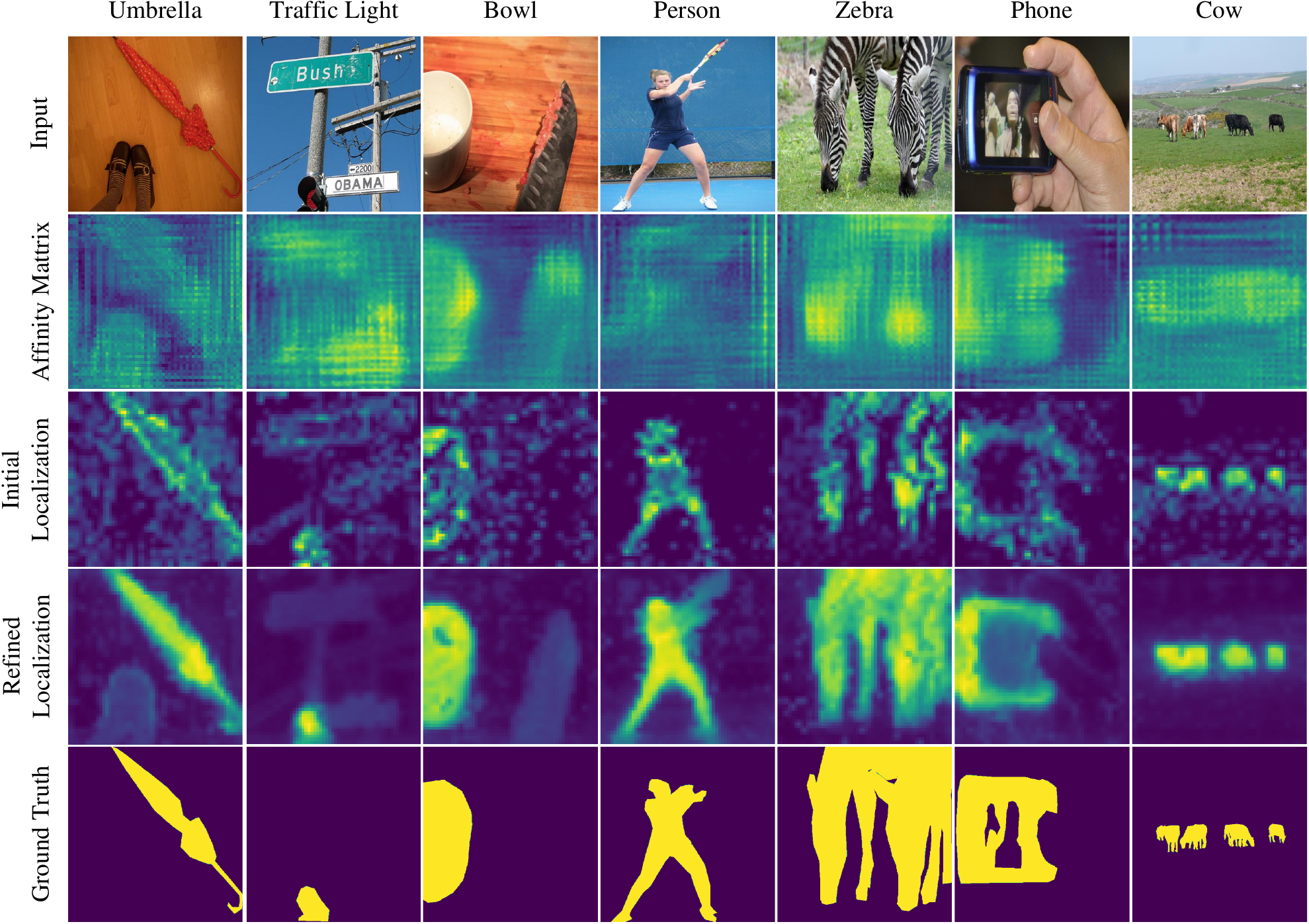}
\caption{Visual comparison of localization based on CLIP. From top to bottom: input image, localization based on affinity matrix, initial localization of our method CLIPix, refined localization of our method CLIPix, ground truth.}
\label{fig5}
\end{figure}

Furthermore, we conducted a more detailed ablation study on the components of the NRC module. As shown in Table \ref{tab5}, we investigated the impact of the noise-resistant correction strategy on the fine localization information $L_{2}$. From the table \ref{tab5}, it can be observed that updating both $E_{new}$ and $C_{new}$ had positive effects, with the best results achieved when both were updated simultaneously. This confirms the rationality and effectiveness of the NRC module. To more intuitively demonstrate the advantages of our method in pixel-level localization, we visualize the localization effects of various types of localization information in Figure \ref{fig5}. It can be observed that the localization information derived directly from computing the affinity matrix between images and texts is rather coarse, failing to provide precise localization and containing a significant amount of noise. In contrast, the initial localization information obtained through our method can effectively capture the key regions of the target object. However, due to CLIP's global bias, the initial localization information still contains some noise. Therefore, our noise-resistant correction strategy optimizes these crucial localization cues, achieving denoising and expanding the localization of the target object region, thereby enhancing the perception of the target object.

\noindent{\bf Effects of Localization Embedding strategy.} Due to the open-set setting, the model has not undergone tuning for unseen classes, resulting in insufficient perception of unseen objects and lacking knowledge of the relationships among object components. Although prior localization information provides localization of partial regions of the target object, the model fails to perceive other regions of the target. To address this limitation, we have designed the Localization Embedding strategy (LE), which retains the original localization information while expanding localization details to accurately predict the entire target object.

The proposed localization embedding strategy integrates localization information with cascaded activation operations, enabling effective expansion from local perception to global perception. It accomplishes the decoding process from local to global through the aggregation of intermediate features. As shown in Table \ref{tab4}, after introducing this strategy, the model achieves significant improvements of {\bf 7.7\%}, {\bf 6.1\%}, and {\bf 8.4\%} in mIoU, FB-IoU, and Boundary-IoU on the PASCAL-5$^{i}$ dataset, respectively. Similarly, on the COCO-20$^{i}$ dataset, the corresponding metrics show notable gains of {\bf 8.8\%}, {\bf 5.0\%}, and {\bf 8.4\%}. This strategy effectively leverages the auxiliary capability of the noise-resistant correction strategy, enabling high-precision perception of category-specific object features based on robust localization information, while appropriately compensating for the lack of edge details to improve edge segmentation performance.

\begin{figure}[t]
\centering
\includegraphics[width=.9\columnwidth]{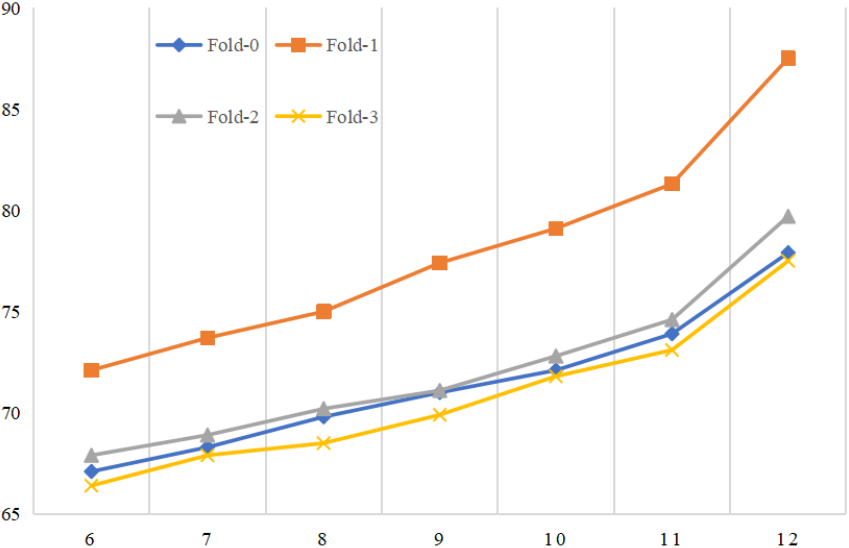}
\caption{Ablation studies of key layer selection on performance impact. The x-axis represents the CLIP layer, and the y-axis represents the performance.}
\label{fig6}
\end{figure}

\noindent{\bf Impact of CLIP key layer selection.} Our proposed method adopts the final layer of CLIP as the key layer and achieves target localization by tracing its attention information during the classification process. To validate the impact of key layer selection on model performance, we present corresponding ablation experiments in Figure \ref{fig6}. Since the final layer of CLIP is closest to the classification output, it exerts the most significant influence on classification results, providing the most representative attention information for localization. As shown in Figure \ref{fig6}, if other layers (non-final layers) are used as the key layer, the segmentation performance drops notably. Experiments also indicate that selecting higher layers closer to the output in CLIP as the key layer further enhances the influence of their attention information on classification results. The localization information provided by such layers is more representative, thereby effectively promoting the improvement of segmentation performance.

\begin{figure}[t]
\centering
\includegraphics[width=.9\columnwidth]{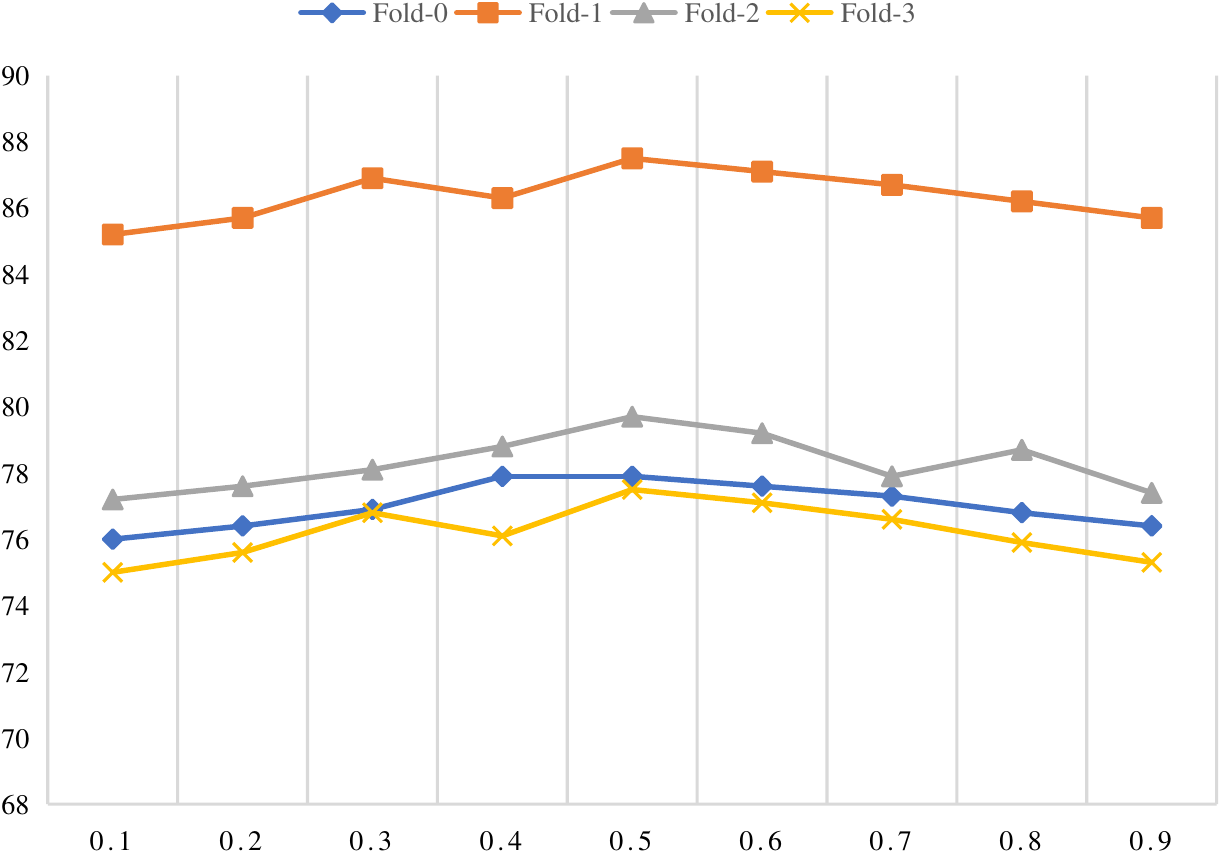}
\caption{Ablation studies on the impact of different parameter settings for $\alpha_{1}$ and $\alpha_{2}$ on performance. The x-axis represents the values of a $\alpha_{1}$, with $\alpha_{2}$ being 1 minus $\alpha_{1}$, and the y-axis displays the mIoU (\%) results.}
\label{fig7}
\end{figure}

\noindent{\bf Influence of the settings of hyperparameters $\alpha_{1}$ and $\alpha_{2}$.} To further validate the effectiveness of our updated class tokens, we conducted ablation experiments about hyperparameters $\alpha_{1}$ and $\alpha_{2}$ in Equation \ref{4} for analysis and verification. As shown in Figure \ref{fig7}, in our experiments on the PASCAL-5$^{i}$ dataset, we varied $\alpha_{2}$ while keeping $\alpha_{1}$ + $\alpha_{2}$ = 1. Performance peaked when $\alpha_{1}$ = $\alpha_{2}$ = 0.5, indicating that both the original class token and the pseudo-target prototype contribute complementary information. Reducing either weight too much leads to performance drops, confirming that both tokens are important.

\subsection{Expansion to multi-class segmentation}
Recently, training-free open-vocabulary segmentation approaches have demonstrated significant application value by enabling multi-class segmentation without the need for training. Given that our noise-resistant correction strategy can provide precise localization information and is inherently training-free, it can be conveniently adapted to this application task to further enhance the practical value of our method. Currently, mainstream methods update the image features of CLIP by reshaping the original attention maps composed of queries (q) and keys (k) to mitigate the impact of global bias for adaptation to pixel-level prediction tasks. However, these attention maps lack strong object-oriented information. To address this, we leverage the precise category localization information from our noise-resistant correction strategy to update the attention maps, thereby integrating object-oriented information. Specifically, based on this localization information, we use a threshold to filter out high-response regions and enhance the correlation between these regions in the attention maps, increasing inter-object correlations to better suit pixel-level prediction tasks. Prior to this, we first utilize CLIP's classification capability to predict the categories present in the image. Then, we combine this with the aforementioned process to sequentially update the category localization information, reducing interference from redundant information and improving inference speed.

We report our experimental results for multi-class segmentation in Table \ref{tab6}. We employ SCLIP \cite{92} as our baseline method. To enhance category-object relevance for dense prediction tasks, we integrate the category localization information obtained through the noise-resistant correction strategy into the attention map of SCLIP. As shown in Table \ref{tab6}, our method significantly improves the baseline performance, achieving a 7.4\% enhancement on the PASCAL VOC 2012 dataset (VOC20). Moreover, notable improvements are still observed on the more challenging PASCAL Context dataset (PC59) \cite{97}, which contains a larger number of categories. 

To further validate the generalization ability of the method, we conducted experiments on more challenging semantic segmentation datasets: Cityscapes (City) \cite{102} and ADE20K (ADE) \cite{103}. Table \ref{tab6} shows that our approach still achieves significant performance improvements, demonstrating that the high‑quality localization information and highly object‑aware features obtained by the model can effectively handle diverse complex scenes. Our method attains state-of-the-art performance on these datasets. Notably, as a training-free approach, it outperforms some trained methods, GroupViT \cite{95} and CLIP-DINOiser \cite{93}, further underscoring the effectiveness and practical value of our proposed method.

\begin{table}[t]
\centering
\caption{Open-vocabulary semantic segmentation quantitative comparison.}
\resizebox{1.0\columnwidth}{!}{
    \begin{threeparttable}
    \begin{tabular}{cc|cccc}    
        \hline
        \textbf{Methods} & \textbf{Train} & \textbf{VOC20} & \textbf{PC59} & \textbf{City} & \textbf{ADE} \\
        \hline
        GroupViT \cite{95}  & $\checkmark$ & 79.7 & 23.4 & 11.1 & 9.2 \\
        TCL \cite{94} & $\checkmark$ & 77.5 & 30.3 & 23.1 & 14.9 \\
        CLIP-DINOiser \cite{93} & $\checkmark$ & 80.9 & 35.9 & 31.7 & 20.0 \\
        \hline
        CLIP \cite{5} & $\times$ & 15.8 & 4.5 & 5.0 & 2.9 \\
        SCLIP \cite{92} & $\times$ & 80.4 & 34.2 & 32.2 & 16.1 \\
        ClearCLIP \cite{91} & $\times$ & 80.9 & 35.9 & 30.0 & 16.7 \\
        ResCLIP \cite{90} & $\times$ & 87.1 & 36.8 & 35.9 & 18.0 \\
        \rowcolor{mygray}\textbf{CLIPix(Ours)} & $\times$ & \textbf{87.8} & \textbf{37.4} & \textbf{37.1} & \textbf{19.1} \\
        \hline
    \end{tabular}
\end{threeparttable}
}
\label{tab6}
\end{table}

\begin{table}[t]
\centering
\caption{Computational efficiency studies on the COCO-20$^{i}$. \enquote{Mem.} refers to memory usage.}
\resizebox{1.0\columnwidth}{!}{
    \begin{threeparttable}
    \begin{tabular}{c|c|cccc}    
        \hline
        \textbf{Methods} & \textbf{Backbone} & \textbf{mIoU}$\uparrow$ & \textbf{FPS}$\uparrow$ & \textbf{FLOPS}$\downarrow$ & \textbf{Mem.}$\downarrow$ \\
        \hline
        LSeg \cite{56} &{ViT-L} & 27.2 & 14 & 320.2G & 9.2G \\
        CLIPSeg \cite{81} & ViT-B & 34.9 & 20 & 210.4G & 6.9G \\
        SAZS \cite{57} & DRN  & 35.2 & 15 & 275.7G & 8.1G \\
        SAZS \cite{57} & ViT-L & 35.3 & 12 & 345.9G & 10.1G \\
        \rowcolor{mygray}\textbf{CLIPix(ours)} & ResNet & \textbf{61.8} & 24 & 140.6G & 4.2G 
        \\
        \rowcolor{mygray}\textbf{CLIPix(ours)} & MobileNet & 57.5 & \textbf{50} & \textbf{33.7G} & \textbf{1.4G}
        \\
        \rowcolor{mygray}\textbf{CLIPix(ours)} & EfficientNet & 59.1 & 38 & 80.2G & 3.1G
        \\
        \hline
    \end{tabular}
\end{threeparttable}
}
\label{tab7}
\end{table}

\subsection{Computational Efficiency Studies}
CLIPix is a lightweight model that exhibits significant efficiency advantages. To validate its lightweight characteristics, we conducted a comparative study on computational efficiency with representative methods on the COCO-20$^{i}$ dataset, with the results presented in Table \ref{tab7}. It can be observed that our model not only substantially outperforms current state-of-the-art methods in terms of performance but also demonstrates clear advantages in computational efficiency—achieving higher FPS with lower computational cost and memory consumption.

This achievement stems from our efficient decoder design. In contrast to the complex decoding structures commonly used in existing studies, our decoder is composed of only a few convolutional and normalization operations, making it lightweight and highly efficient. The design of this lightweight decoder builds upon our efficient encoding pipeline: in the pre-encoding stage, we designed the NRC module, which leverages information attended by CLIP’s key layers during classification to provide precise localization priors for the model; during the encoding stage, we introduce the LE module, which incorporates these localization priors into positional encoding to enhance the model’s high-level perception of target categories and extend its perceptual scope from local to global. Finally, these high-quality image features are fed into the lightweight decoder to generate the final segmentation results. In comparison, complex decoders typically incur substantial computational overhead and increase the risk of overfitting, leading the model to develop a bias toward seen categories and perform poorly on unseen ones.

Edge deployment capability is a key metric for assessing the practical applicability of models. To enhance the suitability of CLIPix for edge devices, we experimented with more lightweight backbone networks such as MobileNetV2 \cite{104} and EfficientNet-B4 \cite{105} for optimization. As shown in Tables \ref{tab1} and \ref{tab7}, our method not only maintains a significant performance advantage but also demonstrates excellent computational efficiency. For real-time AR/VR applications, achieving a frame rate above 90 FPS is typically required to ensure a smooth and stable user experience. However, constrained by the computational capabilities of edge devices and the complexity of the scenes, the actual frame rates in deployment often fall within the range of 30–70 FPS. Although our method outperforms other comparative approaches within the currently acceptable range, it still falls short of the ideal target, necessitating further optimization of computational efficiency in future work.

In subsequent research, we plan to explore the integration of even lighter models, such as CLIP ViT-S, to reduce computational overhead while preserving localization accuracy, and intend to conduct further evaluations using higher-performance GPUs. Additionally, we will design improvement strategies specifically tailored for edge computing scenarios, such as optimizing the attention mechanism for shallow features in CLIP to fully leverage their representational capacity. This approach aims to circumvent redundant inference in deeper network layers, thereby further enhancing overall operational efficiency.

\begin{table*}[t]
\centering 
\caption{Per-class open-set binary semantic segmentation results on COCO-20$^{i}$ in terms of mIoU(\%).}
\label{tab8}
\resizebox{2.0\columnwidth}{!}{
\begin{tabular}{l|cccccccccccccccccccc}
\hline
\textbf{Method} & \textbf{\rotatebox{90}{Person}} & \textbf{\rotatebox{90}{Bicycle}} & \textbf{\rotatebox{90}{Car}} & \textbf{\rotatebox{90}{Motorbike}} & \textbf{\rotatebox{90}{Aeroplane}} & \textbf{\rotatebox{90}{Bus}} & \textbf{\rotatebox{90}{Train}} & \textbf{\rotatebox{90}{Truck}} & \textbf{\rotatebox{90}{Boat}} & \textbf{\rotatebox{90}{Trafficlight}} & \textbf{\rotatebox{90}{Firehydrant}} & \textbf{\rotatebox{90}{Stopsign}} & \textbf{\rotatebox{90}{Parkingmeter}} & \textbf{\rotatebox{90}{Bench}} & \textbf{\rotatebox{90}{Bird}} & \textbf{\rotatebox{90}{Cat}} & \textbf{\rotatebox{90}{Dog}} & \textbf{\rotatebox{90}{Horse}} & \textbf{\rotatebox{90}{Sheep}} & \textbf{\rotatebox{90}{Cow}} \\
\hline
\hline
SAZS &35.7 &56.5 &33.4 &48.2 &74.7 &\textbf{83.2} &16.2 &25.0 &17.6 &13.1 &12.1 &7.3 &56.4 &\textbf{71.9} &12.3 &35.3 &13.8 &17.6 &25.3 &21.1 \\
\rowcolor{mygray}\textbf{Ours} &\textbf{57.9} &\textbf{57.7} &\textbf{68.6} &\textbf{73.0} &\textbf{77.8} &82.4 &\textbf{71.9} &\textbf{73.2} &\textbf{59.2} &\textbf{51.6} &\textbf{83.2} &\textbf{80.3} &\textbf{76.7} &46.2 &\textbf{86.2} &\textbf{80.3} &\textbf{89.0} &\textbf{75.5} &\textbf{82.9} &\textbf{83.0} \\
\hline
\hline
\textbf{Method} & \textbf{\rotatebox{90}{Elephant}} & \textbf{\rotatebox{90}{Bear}} & \textbf{\rotatebox{90}{Zebra}} & \textbf{\rotatebox{90}{Giraffe}} & \textbf{\rotatebox{90}{Backpack}} & \textbf{\rotatebox{90}{Umbrella}} & \textbf{\rotatebox{90}{Handbag}} & \textbf{\rotatebox{90}{Tie}} & \textbf{\rotatebox{90}{Suitcase}} & \textbf{\rotatebox{90}{Frisbee}} & \textbf{\rotatebox{90}{Skis}} & \textbf{\rotatebox{90}{Snowboard}} & \textbf{\rotatebox{90}{Sportsball}} & \textbf{\rotatebox{90}{Kite}} & \textbf{\rotatebox{90}{Baseballbat}} & \textbf{\rotatebox{90}{Baseballglove}} & \textbf{\rotatebox{90}{Skateboard}} & \textbf{\rotatebox{90}{Surfboard}} & \textbf{\rotatebox{90}{Tennisracket}} & \textbf{\rotatebox{90}{Bottle}} \\
\hline
\hline
SAZS &16.2 &57.1 &19.3 &16.5 &\textbf{61.0} &\textbf{78.7} &\textbf{35.5} &\textbf{53.7} &52.1 &43.6 &10.3 &21.1 &\textbf{40.8} &78.5 &19.8 &32.8 &21.4 &18.9 &15.0 &\textbf{69.2} \\
\rowcolor{mygray}\textbf{Ours} &\textbf{90.0} &\textbf{87.4} &\textbf{86.4} &\textbf{85.4} &30.6 &74.6 &34.4 &22.4 &\textbf{76.8} &\textbf{79.4} &\textbf{37.0} &\textbf{56.4} &40.0 &\textbf{61.7} &\textbf{53.6} &\textbf{88.0} &\textbf{31.1} &\textbf{74.5} &\textbf{84.5} &41.1 \\
\hline
\hline
\textbf{Method} & \textbf{\rotatebox{90}{Wineglass}} & \textbf{\rotatebox{90}{Cup}} & \textbf{\rotatebox{90}{Fork}} & \textbf{\rotatebox{90}{Knite}} & \textbf{\rotatebox{90}{Spoon}} & \textbf{\rotatebox{90}{Bowl}} & \textbf{\rotatebox{90}{Banana}} & \textbf{\rotatebox{90}{Apple}} & \textbf{\rotatebox{90}{Sandwich}} & \textbf{\rotatebox{90}{Orange}} & \textbf{\rotatebox{90}{Broccoli}} & \textbf{\rotatebox{90}{Carrot}} & \textbf{\rotatebox{90}{Hotdog}} & \textbf{\rotatebox{90}{Pizza}} & \textbf{\rotatebox{90}{Donut}} & \textbf{\rotatebox{90}{Cake}} & \textbf{\rotatebox{90}{Chair}} & \textbf{\rotatebox{90}{Sofa}} & \textbf{\rotatebox{90}{Pottedplant}} & \textbf{\rotatebox{90}{Bed}} \\
\hline
\hline
SAZS &14.5 &\textbf{58.6} &\textbf{58.9} &39.0 &\textbf{79.3} &\textbf{80.2} &4.8 &10.2 &9.7 &16.0 &4.1 &44.6 &37.4 &60.6 &10.5 &18.1 &36.7 &11.4 &\textbf{46.7} &47.2 \\
\rowcolor{mygray}\textbf{Ours} &\textbf{63.3} &51.7 &16.3 &\textbf{57.1} &28.4 &31.1 &\textbf{72.9} &\textbf{56.4} &\textbf{75.6} &\textbf{63.7} &\textbf{54.9} &\textbf{52.5} &\textbf{87.0} &\textbf{80.8} &\textbf{74.3} &\textbf{69.6} &\textbf{43.5} &\textbf{64.1} &11.6 &\textbf{67.2} \\
\hline
\hline
\textbf{Method} & \textbf{\rotatebox{90}{Diningtable}} & \textbf{\rotatebox{90}{Toilet}} & \textbf{\rotatebox{90}{Tvmonitor}} & \textbf{\rotatebox{90}{Laptop}} & \textbf{\rotatebox{90}{Mouse}} & \textbf{\rotatebox{90}{Remote}} & \textbf{\rotatebox{90}{Keyboard}} & \textbf{\rotatebox{90}{Cellphone}} & \textbf{\rotatebox{90}{Microwave}} & \textbf{\rotatebox{90}{Oven}} & \textbf{\rotatebox{90}{Toaster}} & \textbf{\rotatebox{90}{Sink}} & \textbf{\rotatebox{90}{Refrigerator}} & \textbf{\rotatebox{90}{Book}} & \textbf{\rotatebox{90}{Clock}} & \textbf{\rotatebox{90}{Vase}} & \textbf{\rotatebox{90}{Scissors}} & \textbf{\rotatebox{90}{Teddybear}} & \textbf{\rotatebox{90}{Hairdrier}} & \textbf{\rotatebox{90}{Toothbrush}} \\
\hline
\hline
SAZS &\textbf{43.9} &38.2 &53.1 &\textbf{77.0} &78.9 &\textbf{71.7} &14.0 &16.4 &13.1 &10.8 &6.2 &26.9 &27.9 &40.8 &41.8 &26.8 &41.9 &12.1 &\textbf{29.8} &28.1 \\
\rowcolor{mygray}\textbf{Ours} &23.9 &\textbf{62.6} &\textbf{59.2} &75.9 &\textbf{82.2} &71.0 &\textbf{60.2} &\textbf{81.5} &\textbf{38.2} &\textbf{53.6} &\textbf{69.3} &\textbf{37.9} &\textbf{42.9} &\textbf{56.2} &\textbf{60.1} &\textbf{62.5} &\textbf{62.1} &\textbf{72.7} &0.1 &\textbf{65.9} \\
\hline
\end{tabular}
 }
\end{table*}

\begin{table}[t]
\centering 
\caption{Per-class open-set binary semantic segmentation results on PASCAL-5$^{i}$ in terms of mIoU(\%).}
\label{tab9}
\resizebox{1.0\columnwidth}{!}{
\begin{tabular}{l|cccccccccc}
\hline
\textbf{Method} & \textbf{\rotatebox{90}{Aeroplane}} & \textbf{\rotatebox{90}{Bicycle}} & \textbf{\rotatebox{90}{Bird}} & \textbf{\rotatebox{90}{Boat}} & \textbf{\rotatebox{90}{Bottle}} & \textbf{\rotatebox{90}{Bus}} & \textbf{\rotatebox{90}{Car}} & \textbf{\rotatebox{90}{Cat}} & \textbf{\rotatebox{90}{Chair}} & \textbf{\rotatebox{90}{Cow}} \\
\hline
\hline
SAZS &74.8 &34.9 &83.0 &63.6 &56.9 &78.9 &54.3 &84.0 &20.9 &83.2 \\
\rowcolor{mygray}\textbf{Ours} &\textbf{86.3} &\textbf{48.8} &\textbf{93.8} &\textbf{78.3} &\textbf{82.2} &\textbf{95.4} &\textbf{92.1} &\textbf{96.3} &\textbf{58.4} &\textbf{95.4} \\
\hline
\hline
\textbf{Method} & \textbf{\rotatebox{90}{Diningtable}} & \textbf{\rotatebox{90}{Dog}} & \textbf{\rotatebox{90}{Horse}} & \textbf{\rotatebox{90}{Motorbike}} & \textbf{\rotatebox{90}{Person}} & \textbf{\rotatebox{90}{Pottedplant}} & \textbf{\rotatebox{90}{Sheep}} & \textbf{\rotatebox{90}{Sofa}} & \textbf{\rotatebox{90}{Train}} & \textbf{\rotatebox{90}{Tvmonitor}} \\
\hline
\hline
SAZS &40.5 &81.8 &73.8 &70.1 &37.0 &19.3 &81.8 &44.1 &75.8 &30.1 \\
\rowcolor{mygray}\textbf{Ours} &\textbf{58.8} &\textbf{92.3} &\textbf{90.8} &\textbf{81.1} &\textbf{77.3} &\textbf{74.0} &\textbf{95.6} &\textbf{81.1} &\textbf{85.4} &\textbf{51.7} \\
\hline
\end{tabular}
 }
\end{table}

\begin{figure}[t]
\centering
\includegraphics[width=0.9\linewidth]{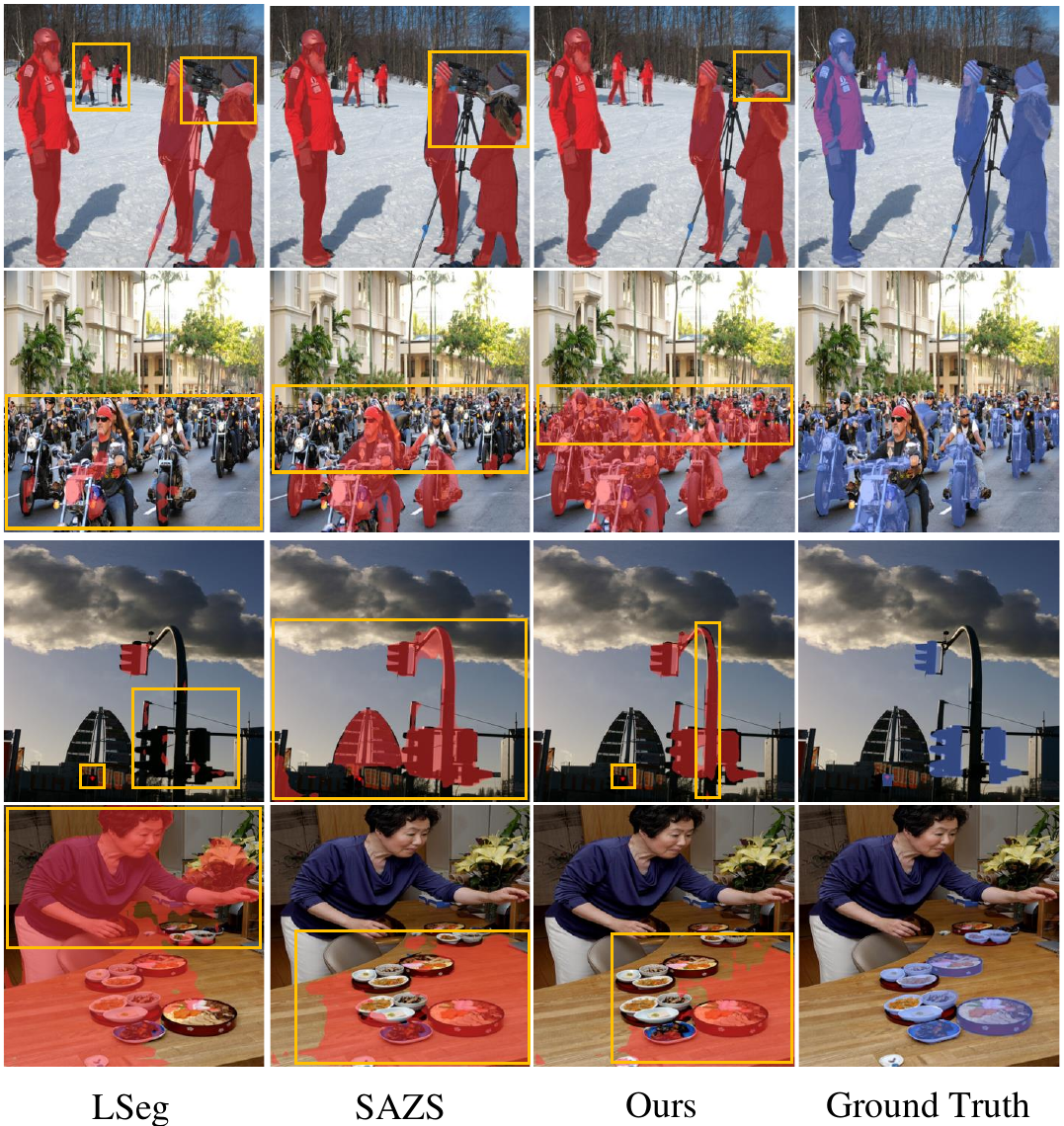}
\caption{Visualized analysis of failure cases in complex scenarios. From left to right: prediction of LSeg, prediction of SAZS, prediction of our method CLIPix, ground truth.}
\label{fig8}
\end{figure}

\subsection{Per-Class Evaluation}
To comprehensively observe the performance of our method, we conducted a detailed evaluation across all classes during the testing phase. As shown in Tables \ref{tab8} and \ref{tab9}, we list the segmentation performance for each specific class in PASCAL-5$^{i}$ and COCO-20$^{i}$, respectively. It can be observed that previous methods exhibited poor performance in classes such as \enquote{bicycle}, \enquote{chair}, and \enquote{person}. This is because these classes often appear alongside other interfering classes, leading to confusion. Especially on the highly challenging COCO-20$^{i}$ dataset, which is full of such difficult classes, each image typically contains around three classes. To our surprise, our method has achieved significant improvements in these challenging classes. This is attributed to our guidance of CLIP, which provides precise localization, and the Noise-Resistant Correction strategy effectively mitigates the interference from objects of other classes.

To further explore potential directions for future method improvements, we present an analysis of failure cases in Figure \ref{fig8}. When there are multiple segmentation targets or interfering objects present (the first and second rows of Figure \ref{fig8}), the model may struggle to achieve precise localization, which is a common challenge in semantic segmentation tasks. Furthermore, we observe that CLIPix performs less ideally in segmenting certain objects (such as forks, spoons, hairdrier, etc.) compared to other categories (such as animals, natural objects, etc.). As shown in Tables \ref{tab8} and \ref{tab9}, we notice an interesting phenomenon that most of these objects with poor segmentation results are instrumental objects created by humans, while the segmentation performance of objects that have naturally evolved or exist in nature (such as animals, plants, etc.) is relatively better. This phenomenon may be related to the following factors:

\noindent{\bf Differences in the distribution of training data for CLIP:} natural objects tend to dominate large-scale image datasets, allowing the model to learn rich feature representations from them. In contrast, the number of images of instrumental objects is relatively smaller. This imbalance in data distribution may lead to inadequate learning by the model for instrumental objects.

\noindent{\bf Essential differences between natural objects and instrumental objects:} the morphological, textural, and structural visual features of naturally evolved objects are repeatedly presented in image data, enabling the model to learn rich feature representations. In contrast, human-created instrumental objects typically have relatively homogeneous visual features and are not formed through natural evolution, which may result in significant differences in their features compared to natural objects.

Indeed, these issues are ubiquitous across many models, and it's not that we completely fail to handle the segmentation of instrumental objects. In fact, we can achieve precise segmentation for certain instrumental objects, such as "tennis rackets", "fire hydrants", "cellphones", and others, even when other models struggle with them. Overcoming this challenge and achieving accurate segmentation for the majority of instrumental objects represents the direction of our future efforts.

\section{Conclusion, Limitations, and Future Work}
In this paper, we introduced CLIPix, a novel framework that repurposes CLIP to achieve precise pixel-level localization for open-set segmentation. By leveraging CLIP’s classification process, CLIPix identifies object-specific attentive regions, using them as localization cues for dense prediction tasks. Our Noise-Resistant Correction strategy refines these cues, reducing noise, while the Localization Embedding strategy enhances detail, enabling accurate segmentation across diverse categories. Extensive experiments on the PASCAL and COCO datasets demonstrate that CLIPix achieves state-of-the-art performance, highlighting its effectiveness and generalization capability.

\textbf{Limitations and Future Work.} While CLIPix significantly improves CLIP's pixel-level localization, it relies on CLIP's pre-existing image-level alignment, which may limit performance in highly complex or cluttered scenes. In future work, we aim to explore adaptive tuning methods to further enhance CLIPix’s robustness in such environments. Furthermore, we will also explore strategies for CLIPix to achieve precise segmentation of instrumental objects.
\bibliographystyle{ieeetr}
\bibliography{TMM}

\end{document}